\documentclass[sn-mathphys,Numbered]{sn-jnl}% Math and Physical Sciences Reference Style
\usepackage[utf8]{inputenc}
\usepackage{graphicx}%
\usepackage{multirow}%
\usepackage{amsmath,amssymb,amsfonts}%
\usepackage{amsthm}%
\usepackage{mathrsfs}%
\usepackage[title]{appendix}%
\usepackage{xcolor}%
\usepackage{textcomp}%
\usepackage{manyfoot}%
\usepackage{booktabs}%
\usepackage{algorithmicx}%
\usepackage{algpseudocode}%
\usepackage{listings}%
\usepackage{dirtytalk}
\usepackage{hyperref}
\usepackage{pifont}% http://ctan.org/pkg/pifont
\newcommand{\xmark}{\ding{55}}%
\usepackage{caption}
\usepackage{tabularx}
\usepackage[ruled,linesnumbered]{algorithm2e}

\usepackage{float}
\usepackage{tikz}
\usepackage{pgfplots}
\pgfplotsset{compat=1.15}
\usepackage{mathrsfs}
\usetikzlibrary{arrows}
\usepackage{tikz-cd}
\usepackage{tkz-graph}
\usetikzlibrary{decorations.pathmorphing,decorations.markings,arrows,calc,shapes.geometric,arrows.meta,positioning}
\tikzset{math3d/.style=
	{x= {(-0.353cm,-0.353cm)}, z={(0cm,1cm)},y={(1cm,0cm)}}}
\tikzset{JLL3d/.style=
	{x= {(0.4cm,-0.2cm)}, z={(0cm,1cm)},y={(-1cm,0cm)}}}
\usepackage{perpage} 
\MakePerPage{footnote}
\usepackage{framed}
\usepackage{makecell}

\usepackage{longtable}
\definecolor{Chocolat}{rgb}{0.36, 0.2, 0.09}
\definecolor{BleuTresFonce}{rgb}{0.215, 0.215, 0.36}
\definecolor{BleuMinuit}{RGB}{0, 51, 102}

\hypersetup{citecolor=BleuMinuit, linkcolor=Chocolat, filecolor=black, urlcolor=BleuMinuit}

\usepackage{xargs} 
\definecolor{armygreen}{rgb}{0.29, 0.33, 0.13}    

\usepackage[colorinlistoftodos,prependcaption,textsize=small]{todonotes}

\newcommandx{\change}[2][1=]{\todo[linecolor=green,backgroundcolor=green!25,bordercolor=green,#1]{#2}}

\newcommandx{\info}[2][1=]{\todo[linecolor=yellow,backgroundcolor=yellow!25,bordercolor=yellow,#1]{#2}}
\newcommandx{\question}[2][1=]{\todo[linecolor=blue,backgroundcolor=blue!25,bordercolor=blue,#1]{#2}}
\newcommandx{\idea}[2][1=]{\todo[linecolor=orange,backgroundcolor=orange!25,bordercolor=orange,#1]{#2}}
\newcommandx{\link}[2][1=]{\todo[linecolor=black,backgroundcolor=white!25,bordercolor=black,#1]{#2}}

\newcommand{\defi}[1]{\emph{#1}}

\newcommandx{\commentST}[2][1=]{\todo[linecolor=cyan,backgroundcolor=blue!25,bordercolor=cyan,#1]{#2}}

\renewcommand{\bf}[1]{\mathbf{#1}}
\def\G_#1{\mathfrak{#1}} 
\def\t_#1{\widetilde{#1}} 
\newcommand{\norm}[1]{\left\lVert#1\right\rVert}

\theoremstyle{thmstyleone}%
\theoremstyle{thmstyletwo}%

\theoremstyle{thmstylethree}%

\begin{document}

\title[Federated learning in wind and solar energy applications: Potential, challenges, and future directions]{A review of federated learning in renewable energy applications: Potential, challenges, and future directions}


\author*[1]{\fnm{Albin} \sur{Grataloup}}\email{albin.grataloup@bfh.ch}

\author*[1,2]{\fnm{Stefan} \sur{Jonas}}\email{stefan.jonas@bfh.ch}

\author[1,3]{\fnm{Angela} \sur{Meyer}}

%% I MADE THE FONT A LITTLE SMALLER HERE BECAUSE THE BLOCK WAS KIND OF BIG TAKING AWAY FROM THE TITLE AND AUTHORS %%%

\affil[1]{\orgdiv{\small{School of Engineering and Computer Science}}, \orgname{Bern University of Applied Sciences}, \orgaddress{\street{Quellgasse 12}, \city{Biel}, \postcode{2501}, \country{Switzerland}}} 

\affil[2]{\orgdiv{\small{Faculty of Informatics}}, \orgname{Università della Svizzera Italiana}, \orgaddress{\street{Via la Santa 1}, \city{Lugano-Viganello}, \postcode{6962}, \country{Switzerland}}}

\affil[3]{\orgdiv{\small{Department of Geoscience and Remote Sensing}}, \orgname{Delft University of Technology}, \orgaddress{\street{Stevinweg 1}, \city{Delft}, \postcode{2628}, \country{Netherlands}}}

\abstract{Federated learning has recently emerged as a privacy-preserving distributed machine learning approach. Federated learning enables collaborative training of multiple clients and entire fleets without sharing the involved training datasets. By preserving data privacy, federated learning has the potential to overcome the lack of data sharing in the renewable energy sector which is inhibiting innovation, research and development. Our paper provides an overview of federated learning in renewable energy applications. We discuss federated learning algorithms and survey their applications and case studies in renewable energy generation and consumption. We also evaluate the potential and the challenges associated with federated learning applied in power and energy contexts. Finally, we outline promising future research directions in federated learning for applications in renewable energy.}


\maketitle
%\tableofcontents

\section{Introduction}  
Renewable energy generation is projected to grow significantly and make up 38\% of the total global electricity production by 2027 \cite{ieaRenewables20222022}. Machine learning models, particularly deep neural networks, have demonstrated remarkable success in improving the operation of renewable energy plants, power grids, and power consuming assets \cite{aslamSurveyDeepLearning2021, MLSustainableReview, wangReviewDeepLearning2019a, izanlooDevelopmentMachineLearning2022}. The models enable data-driven condition monitoring as well as more accurate forecasts of the power production and operational constraints to inform plant operators, traders, and grid managers \cite{stetcoMachineLearningMethods2019}. Machine learning can also improve the real-time control of power plants to ensure optimal responses to changing environmental conditions \cite{wuDeepLearningAdaptive2018}. Training a machine learning model requires a large amount of past operation data \cite{dingCaseStudyAugmentation2019}. An increasing amount of data is being collected to this end. For example, new wind farms may be equipped with sensing and control systems that can generate terabytes of data per day \cite{Gamesa2023}, and modern office buildings can produce hundreds of gigabytes each day \cite{Deloitte2021}. However, many power system assets lack representative data for training asset-specific machine learning models \cite{Kusiak2016}. Newly installed systems completely lack past operation data to train on. Other systems hardly have any representative training data due to prolonged shutdowns or because software and hardware updates altered the systems' operation behavior. \\

A shortage of model training data can, in principle, be mitigated by sharing training data across systems and fleets: For example, a data-scarce wind turbine (WT) may benefit from past operation data of other WTs in the same wind farm or even from the global WT fleet. \\

Renewable energy systems are largely made up of fleets of distributed power assets capable of generating, transmitting, storing, or consuming renewable energy. In addition to WT fleets, they comprise fleets of photovoltaic (PV) power plants, building technology fleets such as heat pumps, or fleets of rechargeable battery storage. We refer to a fleet as the set of all power system assets. Thus, assets especially within a fleet can benefit from data and models trained on data from similar assets of other operators. For example, an operator wants to train fault detection models to detect early-stage damage. No cases of this damage type have been observed yet at her power plant, so she wants to learn from sensor data of other operators who run similar power plants and which experienced that type of damage \cite{TowardsFleet-wideSharingofWindTurbineConditionInformationthroughPrivacy-preservingFederatedLearning}. The transfer of knowledge from another system belongs to the area of transfer learning \cite{panSurveyTransferLearning2010, ComprehensiveSurveyTL}, which has been shown to be effective for neural networks in renewable energy applications \cite{TransferLearningforRenewableEnergySystems:ASurvey}.\\

In practice, however, no data sharing is in place. There is a “lack of data sharing in the renewable-energy industry [which] is hindering technical progress and squandering opportunities for improving the efficiency of energy markets” \cite{Kusiak2016}. The resulting data scarcity inhibits research, innovation and transparency. Operating data of power system assets is usually only accessible to operators, owners, and manufacturers. Many manufacturers seek to maintain control over the data generated by their products and prefer not to make the data accessible as they see business interests at risk \cite{Kusiak2016, agahariItNotOnly2022}. Under these conditions, alternative solutions are needed to enable collaborative learning within and across renewable energy fleets. \\

Federated learning (FL) has been established as a promising solution to address the lack of data sharing. FL has emerged as an approach for distributed parties to collaboratively train machine learning models in a way that preserves the data privacy of all participating parties. FL achieves this by only exchanging and communicating model parameters but not the participants' data. FL has already shown successful results and adoption in multiple application areas, especially in mobile devices \cite{liReviewApplicationsFederated2020}. It has also already been investigated in numerous applications in the context of renewable energy, as presented in this review. Yet, to the best of our knowledge, there has been no survey published of FL applications in the context of renewable energy. While \cite{ChengReview} recently conducted a brief review on its applications in power distribution and transmission systems, it is limited in scope and focused on smart grids. \\

We present a comprehensive review of FL applications in renewable energy generation, consumption, and storage. Our main contributions are: 
\begin{enumerate}[\hspace{1cm}(a)]
    \item an overview of federated learning in renewable energy applications,
    \item a comprehensive literature review, and
    \item an analysis and discussion of the potential, challenges, and promising research directions of FL in renewable energy applications.
\end{enumerate}

The remainder of this review is structured as follows. Section \ref{sec:federated_learning} provides an introduction to federated learning. Section \ref{sec:federated_learning_algorithms} presents the most relevant FL algorithms in more detail. In section \ref{sec:fl_in_renewable_energy_applications}, we provide an overview of FL in renewable energy applications. We discuss its potential, challenges, and possible future research directions in section \ref{sec:potential_challenges_and_futur_directions}. Section \ref{sec:conclusion} presents our conclusions.

\section{Federated learning}
\begin{figure}[H]
    \centering
    %\captionsetup{justification=centering}
    \includegraphics[scale=1.05]{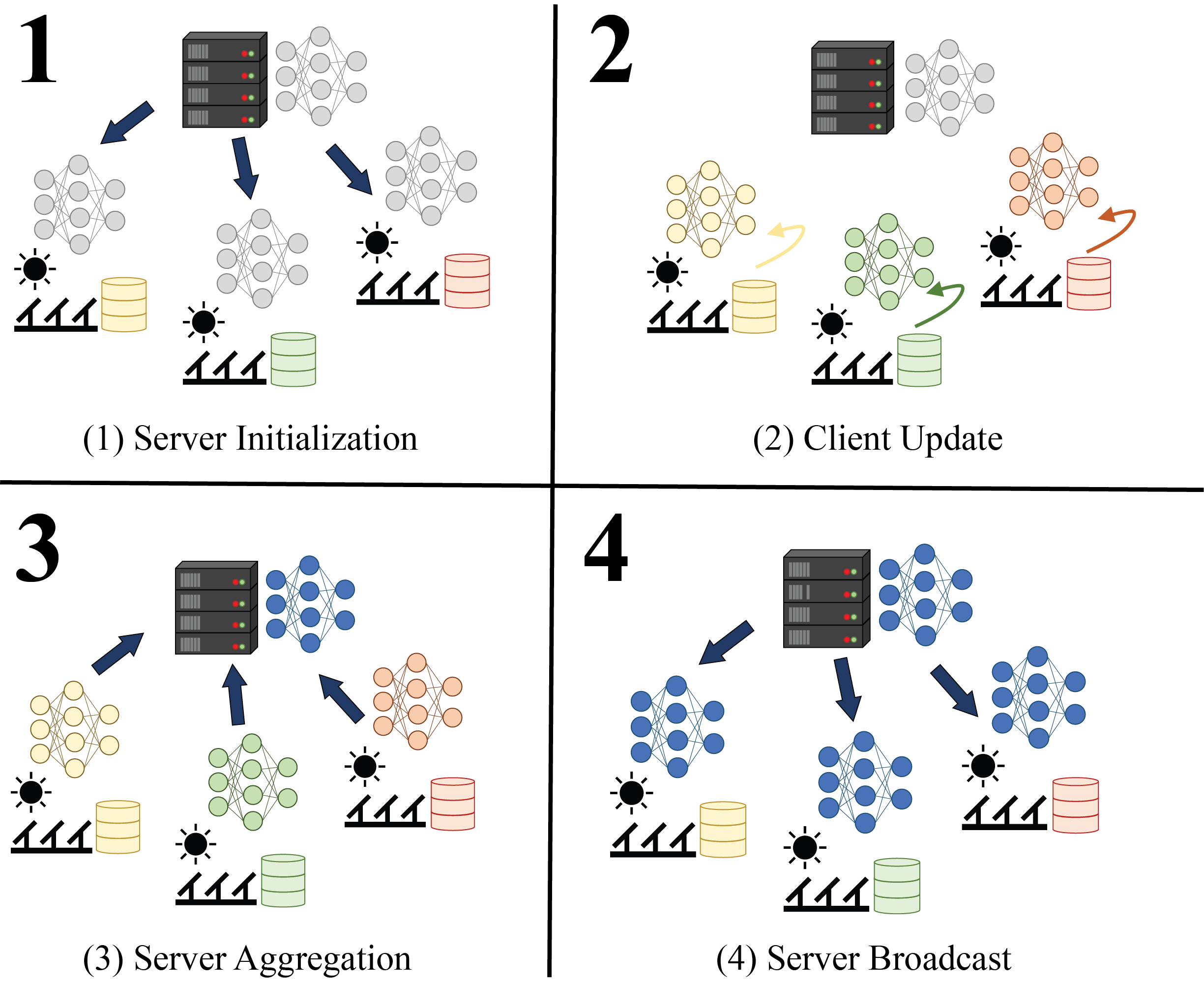} 
    \caption{The general procedure of federated learning, exemplified by PV systems as separate clients with private datasets and a central server (top).}
    \label{fig:fl_steps}
\end{figure}

\
\label{sec:federated_learning}
Federated learning is a technique for collaborative learning of machine learning models by distributed participants referred to as clients. To give an example, a fleet of PV systems, each representing one client, may aim to collaboratively learn a data-driven model for detecting operation faults without exchanging any operation data. The FL training process is iterative and typically involves the clients first locally training a fault detection neural network using only their private dataset, and then transmitting their learned model parameters, i.e., \textit{weights}, to the server. The server then aggregates the received weights according to a specified aggregation algorithm and distributes its computed model state back to the clients. Iteratively, the server receives, aggregates, and distributes weights such that ultimately all stations receive a \textit{global} FL model, containing fault detection knowledge also from other clients without ever sharing any client's operation data. This procedure is illustrated in Figure \ref{fig:fl_steps}. Thus, FL ensures the privacy of locally stored client data and can provide a viable solution to the lack of data sharing. \\

Formally, the setup of federated learning requires the following \cite{FederatedTransferLearning}:
\begin{itemize}
\item[-] A set of $N$ \defi{clients} $C_i$, $i=1\dots N$, participating in the federated learning process.
For instance, $N$ wind turbines.

\item[-] Each client $i$ holds a private dataset $\mathscr{D}_i$ containing $n_i$ samples. 
The datasets remain private, i.e. inaccessible to other clients and the server. In our example $\mathscr{D}_i$ represents the operation data of client WT $C_i$.

\item[-] Data distributions $\pi_{\mathscr{D}_i}$ defining the underlying distribution of each dataset $\mathscr{D}_i$. In our example, $\pi_{\mathscr{D}_i}$ is the distribution that generates the operation data $\mathscr{D}_i$ of WT $C_i$.

\item[-] Client models $\mathscr{M}_i$, $i=1\dots N$, trained on the local dataset $\mathscr{D}_i$. In federated learning, these models are typically deep neural networks parameterized by their weights $\omega_i$.

\end{itemize}

Federated learning involves the following iterative steps:

\begin{figure}[H]
  \centering
  \begin{minipage}{.7\linewidth}
    \begin{algorithm}[H]
      \SetAlgoLined
      {R: number of training rounds.}
      \newline Initialization of the server model\;
      
      \For{$r = 1 \cdots R$}{
        N clients receive a global FL model from the server\;
        
        All clients independently perform training updates on this model using only their local datasets $\mathscr{D}_i$\;
        
        The clients send the parameters of their updated models $\mathscr{M}_i$ to the server\;
        
        The server aggregates all models $\mathscr{M}_i$, $i=1 \dots N$, to obtain the updated global FL model\;
      }
      \caption{Federating learning workflow}
    \end{algorithm}
  \end{minipage}
\end{figure}

where the number of training rounds R can be set, for instance, with a convergence criterion based on validation losses. This general FL framework can be tailored to specific applications, for instance, by introducing a selection of client models $\mathscr{M}_i$ prior to the aggregation step, or by training the global FL model in a synchronous or asynchronous manner. In synchronous learning, the server waits for all clients to finish training before aggregation, while in an asynchronous setting only a subset of $k$ clients are required to have completed their training.\\

The essential parts of the FL framework are the \emph{communication} of the model parameters, the \emph{local training} performed by the clients, and the \emph{aggregation} of the models. The most predominantly used FL framework is based on the \texttt{FederatedAveraging} (\texttt{FedAvg}) algorithm \cite{Communication-EfficientLearningofDeepNetworksfromDecentralizedData} in which the aggregation step consists of averaging the received model weights. We discuss \texttt{FedAvg} in more detail in section \ref{sec:fedsgd_and_fedavg}. Numerous works have since explored further enhancements, e.g., in the security or efficiency of the FL framework. We refer to \cite{FederatedTransferLearning, AdvancesandOpenProblemsinFederatedLearning, ASurveyonFederatedLearningSystems:VisionHypeandRealityforDataPrivacyandProtection, FederatedMachineLearning:ConceptandApplications, FederatedLearning:ChallengesMethodsandFutureDirections} for reviews of FL algorithms.\\

Open challenges of federated learning can be divided into the following categories \cite{FederatedLearning:ChallengesMethodsandFutureDirections, AdvancesandOpenProblemsinFederatedLearning}: Communication and efficiency (section \ref{sec:client_network_and_communication}), security and privacy (section \ref{sec:privacy_preserving_learning}), robustness, fairness, and biases (section \ref{sec:robustness}), and statistical heterogeneity (section \ref{sec:statisical_heterogeneity}). We will discuss these challenges and provide examples which address and aim to mitigate these challenges.

\subsection{Communication and efficiency}
\label{sec:client_network_and_communication}
The client-server communication is an important component in federated learning which may significantly affect the efficiency of the learning process. Depending on the framework and available resources (e.g., computational power, bandwidth), the client-server communication can significantly affect the training speed (e.g., \cite{PrivacyEnhancedEnergyPredictioninSmartBuildingusingFederatedLearning}).\\

\begin{figure}[H]
    \centering
    \captionsetup{justification=centering}
    \includegraphics[scale=0.75]{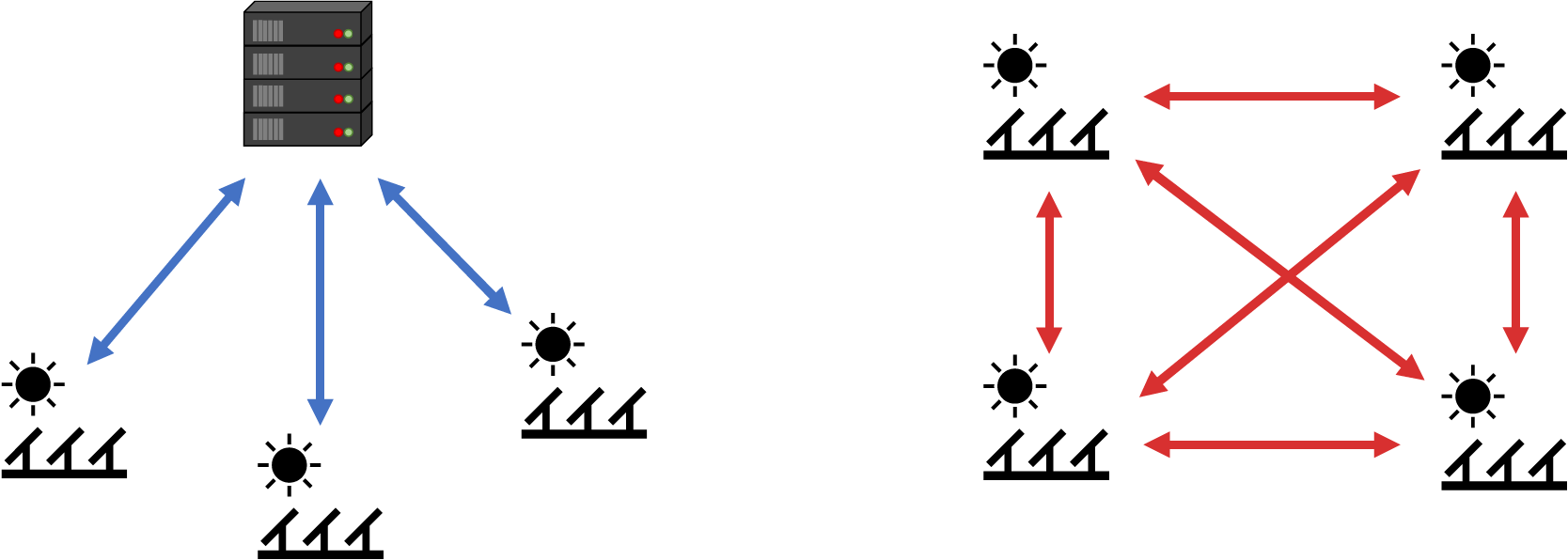} 
    \vspace*{5mm}
    \caption{Centralized (left) vs. decentralized federated learning (right).}
    \label{fig:centralized_vs_decentralized}
\end{figure}

Thus, it may become important to reduce the required number of communication rounds or the overhead of communication. The efficiency can also depend on the communication network topology. We identify two main categories of communication networks, a \emph{centralized} network, in which all weights are sent to a central server for aggregation, and a \emph{decentralized} (i.e., peer to peer) setting where the models are communicated only between clients. These two network types are illustrated in Figure \ref{fig:centralized_vs_decentralized}.\\

There are a variety of options to increase efficiency and to therefore reduce the training time, such as an optimized client selection for each training round, asynchronous peer-to-peer communication, or compression schemes. The appropriate choice may depend on the specific bottleneck, available resources, and further constraints. We refer to \cite{shahidCommunicationEfficiencyFederated2021} and \cite{zhaoEfficientCommunicationsFederated2022} for an overview of these options and a comprehensive survey regarding communication efficiency in federated learning.

\subsection{Privacy and security}
\label{sec:privacy_preserving_learning}
The main motivation for federated learning has been to preserve the data privacy of the clients in collaborative training. However, even information contained in the exchanged model parameters can be maliciously used to obtain sensitive information about client data, thereby violating the privacy guarantees of federated learning. Different kinds of privacy attacks have been presented \cite{FederatedTransferLearning}:

\begin{itemize}
    \item \textbf{Reconstruction attack}: The attacker reconstructs client data from available information, in particular, the communicated model parameters and model updates.
    \item \textbf{Inversion attack}: The attacker tries to reconstruct client data by using the model output.
    \item \textbf{Membership-inference attack}: The goal is to learn if a specific sample is contained inside the training set of the model. The attacker infers whether a sample belongs to the training set or not based on the model output. 
\end{itemize}\

Federated learning is often combined with additional privacy-preserving methods to address these critical vulnerabilities. There are three common classes of defense, namely \emph{secure multi-party computation}, \emph{homomorphic encryption}, and \emph{differential privacy} \cite{FederatedTransferLearning, FederatedLearning:ChallengesMethodsandFutureDirections}. We present the mechanisms of these methods in section \ref{sec:privacy_algos}. While these privacy enhancements assure more data security, they may result in substantial performance and/or efficiency degradation (e.g., \cite{FederatedLearning:ChallengesMethodsandFutureDirections, bagdasaryanDifferentialPrivacyHas2019}).

\subsection{Robustness, fairness, biases}
\label{sec:robustness}
The \textit{robustness} of a model describes how resilient a machine learning model is against adversarial data examples \cite{liDittoFairRobust2020}. Robustness may be achieved by protecting participants from malicious attackers injecting adversarial data samples into the model, ensuring that it does not substantially affect the model performance for the participating clients. Such a situation can also arise unintentionally: For example, a WT client may be affected by an unnoticed sensor fault, causing that client WT to train the local model with faulty samples, which can potentially deteriorate the model performance for other participating client WTs.\\ 

\textit{Fairness} describes how fair a model is to the federated learning participants. Broadly speaking, a model can be described as unfair if some participants are systematically disadvantaged, if there are systematically different performance outcomes, or if clients with similar characteristics obtain disparate results \cite{AdvancesandOpenProblemsinFederatedLearning, liDittoFairRobust2020}. \\

\textit{Biases} inherent in the federated learning process are a major driver in causing unfair circumstances \cite{AdvancesandOpenProblemsinFederatedLearning}. In renewable energy applications, biases may arise from, for example, over-representation of a particular operation mode, a geographical region, or time of day or of year in the training dataset. Biases may also arise from varying computational power available on each client in the FL process, in the selection of clients for each training round which may depend on whether a client is currently online, or in the different amounts of local training data available (which may affect newly installed power systems). Agnostic Federated Learning (AFL) \cite{AgnosticFederatedLearning} provides an approach to reduce biases caused by imbalanced data distributions. We present AFL in section \ref{sec:afl}.\\

We refer to \cite{AdvancesandOpenProblemsinFederatedLearning, zhouFairFederatedLearning2021, lyuPrivacyRobustnessFederated2020} for further mitigation approaches related to robustness, fairness, and biases.

\subsection{Statistical heterogeneity}
\label{sec:statisical_heterogeneity}
Statistical heterogeneity can be a major challenge of federated learning in renewable energy applications. The participating clients of an FL model training process may differ from each other to varying degrees and exhibit individual characteristics in their training datasets. For example, the participating clients can be batteries from different fleets of battery models and vendors, operated under different operating and environmental conditions. If the clients are residential buildings, they may differ by the highly individual energy consumption patterns across different residential buildings. \\

Such variations among participants can lead to differences in the statistical distributions of the clients' local datasets, i.e., to statistical heterogeneity. The data distributions are then referred to as being non-independently and identically distributed (“non-iid”).  Sources of data heterogeneity include (\cite{Federatedlearningonnon-IIDdata:Asurvey}): 

\begin{itemize}
\item \textbf{Label skew}: arises if the distributions of the model's target variables (labels) vary between clients. As an example, label skew may arise if WTs with different power capacities collaboratively learn an FL model of the power output. 
\item \textbf{Feature skew}: arises if the distributions of one or more features differ between the clients. For instance, the mean incoming solar radiation and the operation temperatures of PV systems may differ systematically based on their locations. 
\item \textbf{Quantity skew}: Different clients may have different amounts of training data available. For example, a newly commissioned office building has less historical operation data than buildings that were commissioned previously. 
\item \textbf{Time skew}: The distribution of client data is time dependent. For example, the distribution of wind speed at a WT may depend on the time of year. Based on when the data of a client was acquired, this may bias the client dataset distribution.

\end{itemize}

\subsubsection{Effect on \texttt{FedAvg}}
\label{sec:effect_on_fedavg}
The impacts of non-iid data on the convergence and performance of the \texttt{FedAvg} algorithm are investigated in \cite{SCAFFOLD, hsuMeasuringEffectsNonIdentical2019,OntheConvergenceofFedAvgonNon-IIDData}. Particular attention is devoted to the inferior convergence performance which is attributed to the client drift (\cite{SCAFFOLD, NonIIDDataSilos}): Over the course of FL iterations, each client’s optimal weights $w_i^*$ diverge from the true server optimum $w^*$, thereby leading to poor convergence. The FL algorithm \texttt{FedProx} was introduced in \cite{FederatedOptimizationinHeterogeneousNetworks} to enable a more robust convergence by penalizing large deviations of local weights from the globally averaged weights. We outline the \texttt{FedProx} algorithm in section \ref{sec:fedprox_algo}.

The Federated Uncertainty-Aware Learning Algorithm (\texttt{FUALA}) \cite{FederatedUncertainty-AwareLearningforDistributedHospitalEHRData}, as outlined in section \ref{sec:fuala}, provides an approach to reduce the influence of non-iid data by using a model generalization scoring during the client selection process.

\subsubsection{Customizing FL models to individual clients }
\label{sec:pers_FL}
Under strong non-iid conditions, a global model as obtained by \texttt{FedAvg} may result in moderate or even poor performance \cite{ hsuMeasuringEffectsNonIdentical2019, NonIIDDataSilos}. Models trained only locally (without FL) may outperform a global FL model under such conditions, as demonstrated for wind power plants in \cite{TowardsFleet-wideSharingofWindTurbineConditionInformationthroughPrivacy-preservingFederatedLearning} and discussed in more detail in section \ref{sec:fl_in_renewable_energy_applications}. Therefore, some clients may not have an incentive to participate in the federated learning process (\cite{TowardsFleet-wideSharingofWindTurbineConditionInformationthroughPrivacy-preservingFederatedLearning, AdaptivePFL, SalvagingLocalAdaptation}). To alleviate this issue, personalized federated learning (PFL) \cite{TowardsPFL} has emerged as a technique that creates a customized model for each participant (rather than a single global FL model for all participants) but retains the advantages of collaborative learning. Possible approaches on how to customize ("personalize") models are vast, and examples include client clustering \cite{HierarchicalClusteringBriggs}, personalized model layers \cite{PersonalizationLayers}, meta-learning \cite{PFLHyperNetworks}, or transfer learning-based approaches \cite{FedAvgWithFineTuning}. We refer to \cite{TowardsPFL} and \cite{SurveyPFL} for a comprehensive overview of customization approaches.

While PFL methods usually outperform \texttt{FedAvg} in the presence of non-iid data, there is no current consensus on which PFL algorithm is preferable, as current research suggests that the optimal choice of personalization technique is client-, task- and dataset-specific \cite{ThreeApproachesPFL, PartialModelPers, SalvagingLocalAdaptation, TowardsPFL}. In section \ref{sec:pfl_algos}, we present common approaches used in renewable energy applications, namely \emph{clustering}, \emph{local fine-tuning}, and \emph{personalization layers}. 

\section{Federated learning algorithms}
\label{sec:federated_learning_algorithms}
This section provides an overview of federated learning algorithms that are relevant to renewable energy applications and address the challenges outlined in section \ref{sec:federated_learning}. We discuss model aggregation (section \ref{sec:aggragation_algo}), privacy and security (section \ref{sec:privacy_algos}), and customization algorithms (\textit{personalization}, section \ref{sec:pfl_algos}). Specifically, this includes:

\begin{itemize}
    \item[-] the \texttt{FedProx} alogrithm that can improve the performance of FedAvg on heterogeneous data,
    \item[-]  Agnostic Federated Learning (\texttt{AFL}) to ensure fairness of models,
    \item[-] the Federated Uncertainty-Aware Learning Algorithm (\texttt{FUALA}) that favors clients with models showing better generalization performance,
    \item[-] Privacy enhancements, specifically, secured multi-party computation, homomorphic encryption and differential privacy, and
    \item[-] Personalization methods, namely clustered federated learning, fine-tuning, and personalization layers.
\end{itemize} 

There is a wide variety of further modifications and extensions for federated learning. We refer to \cite{Asystematicliteraturereviewonfederatedlearning:Fromamodelqualityperspective} for a comprehensive review.

\subsection{Aggregation algorithms}
\label{sec:aggragation_algo}
In this section, a selection of federated learning aggregation algorithms is presented. We start by discussing the idea behind \texttt{FedSGD} and \texttt{FedAvg}, the foundational aggregation algorithms for federated learning and discuss some of their limitations. Selected improvements of \texttt{FedAvg} are then presented based on the challenges and limitations which these improvements address.

\subsubsection{\texttt{FedSGD} and \texttt{FedAvg}}\
\label{sec:fedsgd_and_fedavg}
In standard machine learning, training neural networks is performed by minimizing an objective function with the goal of arriving at optimal model weights $\omega$. Typically, a loss function $\ell (\boldsymbol{x}, \boldsymbol{y}; \omega)$ based on dataset features $\boldsymbol{x}$ and labels $\boldsymbol{y}$ is minimized using gradient-based optimization techniques, with the model being parameterized by $\omega$. In standard federated learning, this objective function is extended to $N$ participating clients $C_i$, $i = 1, \dots N$:

\begin{equation}
\min_{\omega} \sum\limits_{i=1}^N \frac{n_i}{n} \ell_i(\boldsymbol{x}_i, \boldsymbol{y}_i; \omega)
\label{eq:fl_objective}
\end{equation}\\

\noindent to minimize the federated model weights  $\omega$ for all clients and their respective datasets $\boldsymbol{x}_i$ and $\boldsymbol{y}_i$, and where $n=\sum_{i}^{N}n_i$ is the total number of all training samples.  \\

\noindent \textbf{Federated Stochastic Gradient Descent} \cite{shokriPrivacyPreservingDeepLearning2015} (\texttt{FedSGD}) was proposed as an aggregation algorithm that minimizes the objective function in equation \ref{eq:fl_objective} by averaging transmitted gradients from clients in each round, with each round consisting of one epoch computed in one batch. That is, in \texttt{FedSGD} all clients compute and transmit their local gradient of the loss $\ell_i$ with respect to the current weights $w^t$ of round $t$, such that the server can perform one stochastic gradient descent step to update the weights for the following round $t + 1$:

\begin{equation}
\omega^{t+1} = \omega^t - \eta \sum\limits_{i=1}^N \frac{n_i}{n} \nabla \ell_i(\omega^t)
\label{eq:fedsgd}
\end{equation}

where $\nabla \ell_i(\omega_i^t)$ denotes the gradient batch averages of client $C_i$ computed over its datasamples, and $\eta$ is the learning rate.\\

\texttt{FedSGD} guarantees that the global model's total loss converges towards a local minimum. However, \texttt{FedSGD} requires an update of the model at each training step, which induces a heavy cost in communication and slows down the training. \\

\noindent  \textbf{\texttt{Federated Averaging}} \cite{Communication-EfficientLearningofDeepNetworksfromDecentralizedData} (\texttt{FedAvg}) is the leading framework used in federated learning, on which many extensions are based. It substantially speeds up \texttt{FedSGD} by performing multiple updates of the client models before averaging them together, and each epoch can be trained using several mini-batches. That is, the clients themselves perform model training updates for multiple batches or epochs before exchanging parameters. Therefore, this algorithm does not average the gradients but directly employs model weights in the aggregation to obtain the updated server weights $\omega^{t+1}$:
\begin{equation}
\omega^{t+1} = \sum\limits_{i=1}^N  \frac{n_i}{n} \omega_i^{t+1}
\label{eq:fedavg}
\end{equation}
where $\omega_i^{t+1}$ are the updated model parameters computed by client $C_i$. \\

This approach results in significantly faster training and reduced communication cost. While there is no guarantee for the convergence of the global model \cite{FederatedTransferLearning}, \texttt{FedAvg} has demonstrated remarkable empirical success. \texttt{FedAvg} has become a predominantly used FL framework for renewable energy applications (section \ref{sec:fl_in_renewable_energy_applications}), consistently demonstrating its advantages in ease-of-use and flexibility. However, the performance of \texttt{FedAvg} can suffer when faced with strong statistical heterogeneity, as the weights of locally trained models may tend to pull the server weights in different directions.

\subsubsection{\texttt{FedProx}}\
\label{sec:fedprox_algo}
When data distributions across clients are non-iid, \texttt{FedAvg} may suffer from poor convergence and performance issues due to the appearing client drift (section \ref{sec:statisical_heterogeneity}). The idea behind  \texttt{FedProx} \cite{FederatedOptimizationinHeterogeneousNetworks} is to train the local client models such that they stay close to the global model. This is achieved by adding a regularization term to each client loss based on the distance between the local client weights, $\omega_i$, and the weights of the global model, $\omega^t$:
\begin{equation}
h_i(\omega_i, \omega^t ) = \ell_i(\omega_i) + \frac{\mu}{2} \norm{\omega_i - \omega^{t}}^2
\label{eq:fedprox}
\end{equation}
\noindent where $\mu$ is the regularization parameter determining the relative importance of the regularization term. The server then aggregates all new parameters $\omega^{t+1}_i$ to obtain the new global model parameters, $\omega^{t+1}$, as in \texttt{FedAvg}. \\

The regularization term forces the local models to stay close to the initial model and therefore reduces the issue of data heterogeneity. However, since the local models are trained to stay close to the global model, this can slow down the convergence process and require more training rounds. Experiments \cite{FederatedOptimizationinHeterogeneousNetworks} have shown an improved testing accuracy on highly heterogeneous data, a slower convergence on iid data compared to \texttt{FedAvg}, and that the regularization parameter $\mu$ can be set dynamically to improve the trade-off between training speed and accuracy. \texttt{FedProx} is a moderately complex extension of \texttt{FedAvg} and retains the high flexibility of \texttt{FedAvg}, allowing combinations with many other FL algorithms.

\subsubsection{Agnostic Federated Learning}
\label{sec:afl}
Agnostic Federated Learning (AFL) \cite{AgnosticFederatedLearning} aims to achieve model fairness (section  \ref{sec:robustness}) by reducing biases due to imbalanced data distributions across clients. This can occur, for example, when there are highly correlated clusters of clients sharing a very similar data distribution. Some clients will then be over-represented in the learned federated model to the detriment of clients exhibiting more unique data distributions. For example, consider learning a control task for clients representing residential buildings, such as heating control. Such a model could be biased towards a group of buildings housing large families (sharing very similar energy consumption patterns) over a group of single households with more individual energy consumption patterns (e.g., night shift workers, mobility workers, travelers). Considering $N$ client datasets $\mathscr{D}_i$ of size $n_i$ based on distributions  $\pi_{\mathscr{D}_i}$, such biases are due to \texttt{FedAvg} learning a target distribution weighted by the relative dataset sizes of all clients:
\begin{equation}
\pi_t = \sum\limits_{i=1}^N \frac{n_i}{n}\pi_{\mathscr{D}_i}
\label{eq:afl_uniform}
\end{equation}
The authors in \cite{AgnosticFederatedLearning} argue that this target distribution does not always reflect the true objective distribution and that it is susceptible to biases. Instead, AFL optimizes for any possible target distribution formed by an optimal mixture of client data distributions:
\begin{equation}
\pi_\lambda = \sum\limits_{i=1}^N \lambda_i \pi_{\mathscr{D}_i}
\label{eq:afl_mixture}
\end{equation}

 where $\lambda_i$ defines the mixture weight and replaces the standard dataset size weights. Thus, the server optimizes the received weights according to an objective based on equation \ref{eq:afl_mixture} in the aggregation step. AFL aims to ensure that the global model is working well for all clients and not only for a majority of similar clients..

\subsubsection{Model selection and the Federated Uncertainty-Aware Learning Algorithm}
\label{sec:fuala}
Model generalization is an important performance measure. In federated learning, a local model trained on a client’s dataset that generalizes well to multiple other clients can be considered superior to a model only performing well on its own training dataset, as it is more likely to perform well on outliers and previously unseen data. Federated Uncertainty-Aware Learning Algorithm (\texttt{FUALA}, \cite{FederatedUncertainty-AwareLearningforDistributedHospitalEHRData}) is an FL algorithm based on \texttt{FedAvg} that employs model selection to favor client models which generalize well to other clients. This is achieved by selecting those models for aggregation more frequently (i.e., with a higher probability). \\

In \texttt{FUALA}, the client selection at the start of each aggregation round follows a dynamic distribution $\Pi$, first initialized as the uniform distribution. Then, in each round, the distribution is updated as follows:
\begin{figure}[H]
  \centering
  \begin{minipage}{.7\linewidth}
    \begin{algorithm}[H]
      \SetAlgoLined
      \KwData{N clients $C_i, i=1, \dots , N$ with their local datasets $\mathscr{D}_i$ and models $\mathscr{M}_i$.}
    
        The server creates a random permutation of the client indices $\sigma$ and sends $\mathscr{M}_{\sigma_i}$ to $C_i$. \newline (That is, each client receives a model $\mathscr{M}_{j \neq i}$ trained on another client's dataset $\mathscr{D}_j$, in addition to the global model)\;
        
        Each client $C_i$ computes the generalization score (e.g., accuracy) of $\mathscr{M}_{\sigma_i}$ on its own dataset $\mathscr{D}_i$ \newline 
        The client's dataset $\mathscr{D}_i$ acts as a test set for model $\mathscr{M}_{\sigma_i}$\;
        
        The server updates the distribution $\Pi$ such that models with better generalization scores are selected with a higher probability for the next FL round\;
                
      \caption{Distribution update}
    \end{algorithm}
  \end{minipage}
\end{figure}

FUALA only affects the client selection at the start of each FL round. This approach assigns more weight to clients whose models adapt well across the other clients. \texttt{FUALA} was found \cite{FederatedUncertainty-AwareLearningforDistributedHospitalEHRData} to achieve a superior performance compared to other federated learning algorithms.

\subsection{Privacy and security}
\label{sec:privacy_algos}
Privacy enhancements can be added to a standard federated learning framework to ensure stronger privacy by reducing the opportunities of malicious attacks (section \ref{sec:privacy_preserving_learning}). Privacy enhancements include:\\

\textbf{Secure multi-party computation} \cite{bonawitzPracticalSecureAggregation2017} enables the secure computation of functions (e.g., sums) on data from different clients without actually sharing the full data samples. A main approach is secret sharing, based on splitting the transmitted data of each client in shares. These shares can be exchanged between clients without revealing full information of the data, as all computing operations are only performed on the shares of each client. Results computed on each share can then be aggregated to obtain the equivalent result of computing the same operation on all (non-split) data. This mechanism can be applied to aggregate the model parameters without requiring clients to explicitly share their model parameters with the server.\\

\textbf{Homomorphic encryption} \cite{acarSurveyHomomorphicEncryption2019} is an encryption scheme which allows to carry out operations directly over encrypted data. It ensures that an operation, such as multiplication, of two weights $w_1$, $w_2$ can be performed fully within the encrypted space: $HE(w_1 \cdot w_2) = HE(w_1) \cdot HE(w_2)$, where $HE()$ represents homomorphic encryption. In the context of federated learning, this property is useful to defend against reconstruction attacks by encrypting all communicated model parameters and/or gradients, while still allowing aggregation operations in the encrypted space without the need of decryption.\\

\textbf{Differential privacy} \cite{OurDataOurselves:PrivacyViaDistributedNoiseGeneration} provides a security guarantee enabling calculating operations over a dataset while restricting the information that can be retrieved about a specific dataset sample. This property is achieved by adding noise to datapoints (i.e., in FL for instance to data samples or transmitted weights of clients) such as Laplace or Gaussian noise. The extent of this noise is flexible, allowing for a trade-off between privacy guarantees and accuracy. We refer to \cite{TheAlgorithmicFoundationsofDifferentialPrivacy} for an overview of differential privacy. \\

While all these privacy mechanisms can be added to a FL framework to ensure stronger privacy guarantees by mitigating privacy attacks, they do have drawbacks. Secure multi-party computation and homomorphic encryption can substantially affect the efficiency by increasing the communication overhead, while differential privacy can be significantly detrimental to the model accuracy \cite{ouadrhiriDifferentialPrivacyDeep2022}.

\subsection{Personalized federated learning}
\label{sec:pfl_algos}
\begin{figure}[H]

    \centering
    \captionsetup{justification=justified}
    \includegraphics[width=\textwidth]{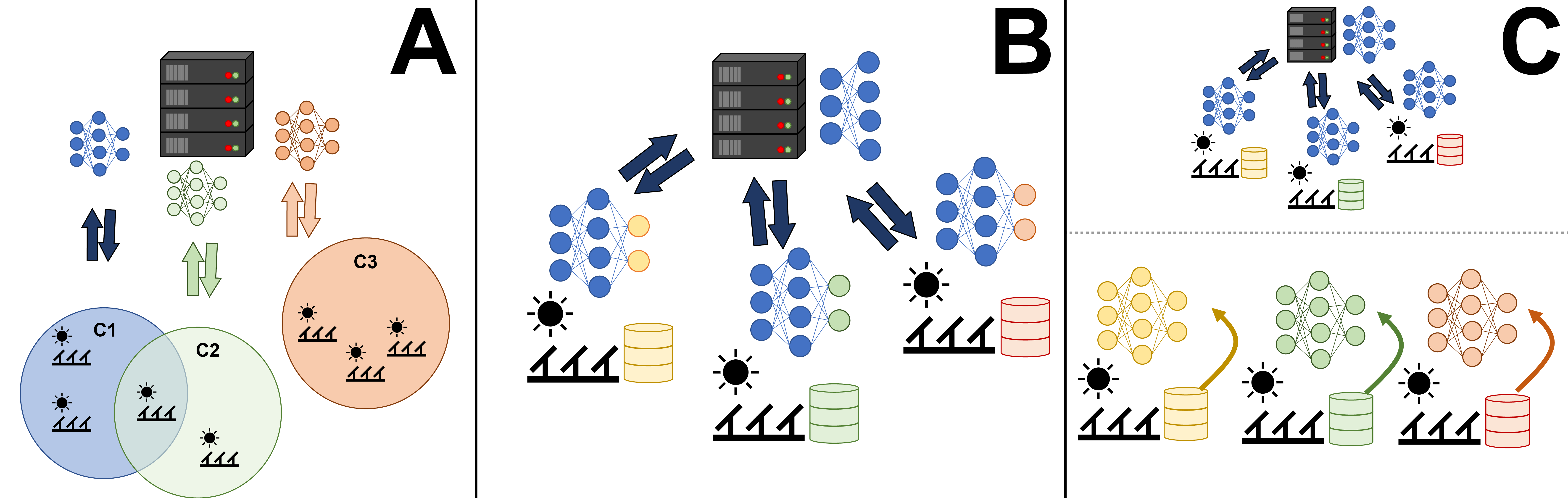}
    \caption{Personalization methods in centralized FL with clients exemplified as PV systems. \textbf{A}: Clustering.  \textbf{B}: Personalization layers. \textbf{C}: Fine-tuning.}
    \label{fig:pers_methods}
\end{figure}

Personalized FL aims to learn models specifically tailored to individual clients (or to a subset of clients) rather than a single, potentially poorly generalizing model. Common personalization techniques include clustering, personalization layers, and fine-tuning, as illustrated in Figure \ref{fig:pers_methods}.

\subsubsection{Clustered federated learning}\
The idea of clustering-based FL approaches is to build multiple separate federated models (e.g., using \texttt{FedAvg}), where each model is trained only on a specific subset of clients, called a cluster. These clusters are selected so that the clients belonging to the same cluster are similar to each other. This reduces the distribution heterogeneity within a cluster with the aim of improving the federated learning performance. Various approaches to clustering clients exist. Clustering can be applied using meta-data of the clients (e.g., by geographic location or power capacity) or by clustering depending on model information (e.g., the distance of the local model weights to the cluster’s model weights) \cite{Multi-CenterFederatedLearning:ClientsClusteringforBetterPersonalization}. \\

This strict type of clustering can give rise to disadvantages. For example, there is a risk of data shortage within small clusters or too rigid clusters focusing only on part of the relevant information (for instance, a cluster containing only newly installed systems might be subject to time skew) without benefiting from knowledge of other relevant clients. To mitigate this issue, some clustering algorithms propose incorporating parameters of clients from other clusters \cite{Privacy-preservingknowledgesharingforfewshotbuildingenergyprediction:Afederatedlearningapproach}. This way, all clusters remain involved in the collaborative training. Fuzzy clustering \cite{ObjectiveFunctionClustering} extends this idea by defining that every client belongs to every cluster to a certain degree, and therefore should be involved in the training of that cluster in proportion to that degree. The fuzzy clustering framework (illustrated in Figure \ref{fig:pers_methods}A) is a promising approach to reduce data heterogeneity while still learning from all clients.

\subsubsection{Personalization layers}
The layers of a neural network can exhibit specific functions depending on their position within the model. For instance, the first layers commonly extract low-level and typically generalizable features, while the top layers (closer to the output) compute highly complex dataset- and task-specific features \cite{ TransferableFeatures}. In multi-task and transfer learning, the knowledge contained in these generalizable and shared representations of selected layers plays a significant role \cite{TransferableFeatures}. This is a main motivation for using personalization layers in federated learning. \\

By keeping weights of selected layers local, i.e., client-specific, and training the remaining layers collaboratively, the resulting customized models can capture general shared information as well as client-specific information. Neural networks can be split into base layers and \textit{personalization layers} to this end \cite{PersonalizationLayers}. The weights of the base layers (the first few general layers) are shared and aggregated as usual through e.g., \texttt{FedAvg}, while the personalized layers (upper layers) remain completely local and private. This approach is illustrated in Figure \ref{fig:pers_methods}B and enables customizing ("personalizing") an FL model for each client participating in the training. This idea can be further generalized by selectively choosing any specific part of the network to be personalized, i.e., any layer or only selected parts of more complex architectures (e.g., sub-components of a transformer) \cite{PartialModelPers, ThinkLocallyActGlobally}.

\subsubsection{Personalized federated learning with fine-tuning}
Transfer learning \cite{panSurveyTransferLearning2010, ComprehensiveSurveyTL} aims to transfer knowledge from a task learnt on a domain in which data and information are abundantly available (“source domain”), to the same or a related task in a domain where typically only few (labelled) data points are available (“target domain”). Fine-tuning is a simple, yet common and successful transfer learning strategy. It involves using a pre-trained model learned on the source domain data and then fine-tuning some or all of its layers by training them on the target domain dataset for a few additional steps \cite{TransferableFeatures, CNNFineTuning, RethinkingHyperparametersFinetuning}. The fine-tuned model will retain transferable and generalizable knowledge from the source domain while being adapted to the specific target dataset characteristics. This approach usually results in improved performance compared to a model trained from scratch on the (scarce) target data only (see e.g., \cite{CNNFineTuning}).\\

In a personalized FL setting, fine-tuning involves first collaboratively training a global model, such as \texttt{FedAvg}. In a second phase, all clients separately fine-tune the global model by training some of its parameters for a few additional steps on only their own local datasets \cite{EvaluationOndevicePersonalization, FederatedAsymptoticsCompare, FedAvgWithFineTuning}. The fine-tuned models still contain generalizable knowledge obtained from other participating clients, but specifically adapted to the client’s local dataset, which reduces the performance degradation caused by statistical heterogeneity. A typical fine-tuning strategy is illustrated in Figure \ref{fig:pers_methods}C.

\section{FL in renewable energy applications}
\label{sec:fl_in_renewable_energy_applications}

\subsection{Resource, production, and load forecasting}

Production forecasts are critical in a renewable energy system for integrating variable energy sources such as solar and wind energy. Typically, a substantial amount of representative training data is required to train a power forecast model, which if lacking might result in inferior model performance. In forecasting, the collected data of a specific site (e.g., a wind farm, PV system, or a residential building) might not be fully representative, for instance due to a limited number of observations which were recorded only during specific weather conditions. The goal of federated learning is to improve the performance of machine learning models which were trained using only local client data, that is, \emph{local models}, while preserving the privacy of these data from other clients. \\

\noindent \textbf{Wind power forecasting}. In wind power forecasting, the clients, i.e., wind turbines or wind farms, aim to collaboratively learn to directly forecast the generated power or wind speeds. The studies in this field \cite{Windpowerforecastingconsideringdataprivacyprotection, DeepFederatedLearning-BasedPrivacy-PreservingWindPowerForecasting, ACyber-Securegeneralizedsupermodelforwindpowerforecastingbasedondeepfederatedlearningandimageprocessing, APrivacy-preservingWindSpeedPredictionMethodBasedonFederatedDeepLearning, AnefficientfederatedtransferlearningframeworkforcollaborativemonitoringofwindturbinesinIoE-enabledwindfarms}  consistently report that FL can be employed to obtain significantly more accurate forecasts compared to local models, demonstrating the potential and benefits of FL. In all these cases, the FL performance also achieves performance comparable to models trained in (privacy-violating) data sharing settings, i.e.,  \emph{centralized models}.

The majority of studies are limited to unmodified \texttt{FedAvg} algorithms, while \texttt{FedProx} is employed in \cite{APrivacy-preservingWindSpeedPredictionMethodBasedonFederatedDeepLearning}, and an adversarial deep domain adaptation framework is used with \texttt{FedAvg} in \cite{AnefficientfederatedtransferlearningframeworkforcollaborativemonitoringofwindturbinesinIoE-enabledwindfarms}.

Improvements to privacy and security are presented in \cite{APrivacy-preservingWindSpeedPredictionMethodBasedonFederatedDeepLearning}  by encrypting the model parameters. In \cite{ACyber-Securegeneralizedsupermodelforwindpowerforecastingbasedondeepfederatedlearningandimageprocessing}, the authors focus on enhancing the security by reducing the potential for data injection attacks.

While significantly less common, another application of FL has been to create a highly generalized model with the main purpose of alleviating the cold-start problem in machine learning. The cold-start problem occurs when there is insufficient historical data to train a model \cite{forootaniTransferLearningbasedFramework2023}. This is a likely occurrence in renewable energy applications, for instance for newly installed power grid infrastructure which completely lacks representative data for training a machine learning model for condition monitoring. By providing a global FL model to these affected clients at the start of their operations, they can still benefit from an adequate task performance despite data scarcity issues. The authors of \cite{ACyber-Securegeneralizedsupermodelforwindpowerforecastingbasedondeepfederatedlearningandimageprocessing} focus mainly on developing such a generalizable model for wind power forecasting, investigating the global FL model performance on wind farms that did not initially participate in the learning process, thereby demonstrating another promising application and benefit of FL.
\newline \newline

\noindent \textbf{Solar forecasting.} In solar power forecasting, clients can be PV systems, collaboratively learning the expected power generation and related tasks. A standard \texttt{FedAvg} approach is employed in \cite{NovelPVPowerHybridPredictionModelBasedonFLCo-TrainingMethod}, showing a highly significant reduction in forecast error compared to models trained only locally on each PV system. The authors further observe a remarkable generalization ability of the learnt global model. Results in \cite{zhangProbabilisticSolarIrradiation2021} show that a \texttt{FedAvg} scheme outperforms local models, achieving comparable results to a centralized model. However, a standard
\texttt{FedAvg} model (e.g., without personalization) does not necessarily always outperform local models. Indeed, \cite{Asolarforecastingframeworkbasedonfederatedlearninganddistributedcomputing, FuzzyClusteredFederatedLearningAlgorithmforSolarPowerGenerationForecasting, APrivacy-PreservingFederatedLearningMethodforProbabilisticCommunity-LevelBehind-the-MeterSolarGenerationDisaggregation} report worse performance with a standard \texttt{FedAvg} scheme compared to local models. This occurrence is to be attributed to statistical heterogeneity, caused for instance by data sources based on substantially different geographical locations (e.g., \cite{Asolarforecastingframeworkbasedonfederatedlearninganddistributedcomputing}). Personalized FL has been employed to address this issue. An approach based on personalization layers for community-level disaggregation is presented in \cite{APrivacy-PreservingFederatedLearningMethodforProbabilisticCommunity-LevelBehind-the-MeterSolarGenerationDisaggregation}, where specific layers of the model remain community-specific and are not shared with the server. An encoder-decoder structure with a two-stage training strategy is presented in \cite{FuzzyClusteredFederatedLearningAlgorithmforSolarPowerGenerationForecasting}, in which only the decoder (forecaster) weights are shared and aggregated, while the encoder (feature extractor) remains local, i.e., customized. In \cite{FuzzyClusteredFederatedLearningAlgorithmforSolarPowerGenerationForecasting}, fuzzy clustering is chosen as personalization method, where clusters are based on meta-information (location). These personalization techniques do not only manage to outperform local models and the standard FL model, but also a centralized model.

Apart from these changes in personalization, the aggregation algorithm employed in previous studies is unmodified \texttt{FedAvg}. Privacy enhancements are implemented in \cite{zhangProbabilisticSolarIrradiation2021}, through a secure aggregation with differential privacy. In terms of communication and efficiency, both \cite{Asolarforecastingframeworkbasedonfederatedlearninganddistributedcomputing} and \cite{APrivacy-PreservingFederatedLearningMethodforProbabilisticCommunity-LevelBehind-the-MeterSolarGenerationDisaggregation} remark uncertainties regarding the training time and efficiency, i.e., possible issues when faced with a significantly larger number of participants.
\newline \newline

\noindent \textbf{Load forecasting.} Federated learning has also been employed to forecast the energy consumption
of buildings. A standard \texttt{FedAvg} approach is shown in \cite{ImprovingtheAccuracyofLoadForecastingforCampusBuildingsBasedonFederatedLearning} to lead to more accurate load forecasting, resulting in superior performance in practically all client buildings compared to local models, as well as comparable results to the centralized setting. However, statistical heterogeneity is a re-occurring theme in load forecasting (for instance caused by highly individual energy consumption patterns), which negatively impacts the performance of standard FL approaches. This issue is extensively investigated in \cite{FederatedLearningforShort-TermResidentialLoadForecasting}, where even the centralized model results in inferior residential load forecasting accuracy compared to local models. The authors first investigate a clustering approach where client updates are used as proxy for client similarity, which however still proves to be inferior, despite improving the non-clustered approach. Only by incorporating a fine-tuning step as personalization method does the proposed FL method result in the most accurate model. 

Similarly, a \texttt{FedAvg} framework enhanced with clustering based on client validation losses with a subsequent fine-tuning is proposed in \cite{Privacy-preservingknowledgesharingforfewshotbuildingenergyprediction:Afederatedlearningapproach} and shown to be the favorable method in comparison to a localized and centralized setting. Another two-stage personalization approach is shown in \cite{PersonalizedFederatedDARTSforElectricityLoadForecastingofIndividualBuildings}, where buildings are first clustered according to model architectures with subsequent local fine-tuning of the federated cluster model. This two-stage strategy is shown to result in improved performance over either only clustering or fine-tuning and in significant improvements compared to an approach without personalization. 

Fine-tuning is also applied in \cite{Federatedlearning-basedshort-termbuildingenergyconsumptionpredictionmethodforsolvingthedatasilosproblem}. A significant benefit of personalization is further shown in \cite{ImpactofFederatedLearningOnSmartBuildings}, and also in \cite{DeepFederatedAdaptation:AnAdaptativeResidentialLoadForecastingApproachwithFederatedLearning}, where the authors propose a method based on domain adaptation.

Regarding privacy and security, a wide variety of solutions is proposed. Parameter encryption \cite{ASecureFederatedDeepLearning-BasedApproachforHeatingLoadDemandForecastinginBuildingEnvironment}, a secure aggregation algorithm based on multi-party security \cite{Federatedlearning-basedshort-termbuildingenergyconsumptionpredictionmethodforsolvingthedatasilosproblem}, and a private data aggregation scheme \cite{Privacy-preservingknowledgesharingforfewshotbuildingenergyprediction:Afederatedlearningapproach} have been presented.

The benefits of the generalization properties of FL in alleviating the cold-start issue are investigated in \cite{Federatedlearning-basedshort-termbuildingenergyconsumptionpredictionmethodforsolvingthedatasilosproblem} showing improvements when only limited data is available, in \cite{ASecureFederatedDeepLearning-BasedApproachforHeatingLoadDemandForecastinginBuildingEnvironment}, suggesting that FL models perform well on out-of-sample distributions, and in \cite{Privacy-preservingknowledgesharingforfewshotbuildingenergyprediction:Afederatedlearningapproach}, further demonstrating good generalizability to nonparticipating buildings. 

Despite these extensions, \texttt{FedAvg} remains the predominant aggregation algorithm, with some variation in \cite{Anadaptivefederatedlearningsystemforcommunitybuildingenergyloadforecastingandanomalyprediction}, where adaptive weights based on client participation are proposed to mitigate the impacts of asynchronous updates. A standard central server structure without efficiency enhancements is present in all but \cite{DecentralizedFederatedLearningFrameworkfortheNeighborhood:ACaseStudyonResidentialBuildingLoadForecasting}, where a substantial increase in efficiency is achieved by a decentralized approach, and in \cite{ImpactofFederatedLearningOnSmartBuildings} using peer-to-peer networks.

%%% ------  FAULT DETECTION AND DIAGNOSTICS -----

\subsection{Fault detection and diagnostics}
Fault detection and diagnostics is a central task to ensure energy plant uptimes, to reduce maintenance costs, and overall to increase the reliability and cost-efficiency of renewable energy systems. However, an extensive and representative fault database is required to detect and classify faults. Detecting anomalous operation behavior in terms of deviations from the normal state also requires a large, representative dataset of past normal operation data. These requirements cannot always be met in renewable energy applications. For example, a database from a wind turbine with a history of blade damages cannot be adequately used to build a fault detection model for gearbox damages. Data from other WTs is usually needed. With federated learning, the participants in the FL training process can share and obtain exposure to a larger number of fault types or normal operation data, without having to share any data from their systems. FL studies of fault detection and diagnostics applications follow similar principles as in forecasting, with the main goal being to achieve superior performance compared to local models. 
\newline \newline

\noindent \textbf{Fault detection for wind turbines.} The FL clients, typically single wind turbines,
collaboratively learn fault detection or classification tasks. An early study \cite{DeepFedWT:Afederateddeeplearningframeworkforfaultdetectionofwindturbines} investigated an unmodified \texttt{FedAvg} scheme for fault detection with two wind turbines of the same type and manufacturer as clients, demonstrating that the FL model can successfully surpass the ability of local models to distinguish between a fault-free and anomalous state of the turbines, reaching comparable results as in a centralized setting.

The authors mention the possibility of non-iid issues, especially if different turbine types were to be considered. This statistical heterogeneity is a theme in \cite{TowardsFleet-wideSharingofWindTurbineConditionInformationthroughPrivacy-preservingFederatedLearning}, where learning normal behavior models of wind turbines of the same type in a fleet is investigated. While a standard \texttt{FedAvg} approach results in more accurate normal behavior models for WT fleet members lacking representative training data, a reduction in accuracy is shown for non-affected WTs. Fine-tuning is introduced both to improve performance and to incentivize participation from fleet members that do not lack data, by guaranteeing them a model performance that is at least equal to that of locally trained models. In terms of efficiency, the authors report a significant increase in required training time with FL. A possible solution to alleviate this issue is demonstrated with a different communication structure approach in \cite{FederatedMulti-ModelTransferLearningBasedFaultDiagnosiswithPeer-to-PeerNetworkforWindTurbineCluster}, where a peer-to-peer network is shown to be more effective. 

Various studies \cite{Windturbinebladeicingdetection:afederatedlearningapproach, AClass-ImbalancedHeterogeneousFederatedLearningModelforDetectingIcingonWindTurbineBlades, HumanKnowledge-basedCompressedFederatedLearningModelforWindTurbineBladeIcingDetection, ABlockchain-EmpoweredCluster-BasedFederatedLearningModelforBladeIcingEstimationonIoT-EnabledWindTurbine} have demonstrated the successful application of FL particularly for blade icing detection. Notably, \cite{Windturbinebladeicingdetection:afederatedlearningapproach} puts emphasis on data imbalance and homomorphic encryption, while in \cite{AClass-ImbalancedHeterogeneousFederatedLearningModelforDetectingIcingonWindTurbineBlades} statistical heterogeneity is addressed with transfer learning-based approaches in which the feature maps themselves (instead of weights) are securely transmitted to the server, showing a significant improvement compared to \texttt{FedAvg}.
\newline \newline

\noindent \textbf{Solar energy.} With FL, PV systems can collaboratively learn to detect or classify
faulty or anomalous behavior. In \cite{AsynchronousDecentralizedFederatedLearningforCollaborativeFaultDiagnosisofPVStations}, an asynchronous decentralized (serverless) FL fault detection framework is proposed. The authors first show that local models fail, as expected, when faced with fault types not present in local datasets. With the proposed FL method, the accuracy reaches that of centralized learning, capable of detecting a wider variety of faults. The authors further report a significant reduction in required training time and number of communication transmissions through their proposed asynchronous decentralized framework. Another decentralized approach \cite{CollaborativelyIGBTOpencircuitFaultsinPhotovoltaicInverters:ADecentralizedFederatedLearning-basedMethod} supports the finding that PV systems can detect faults which never occurred locally, even when fault types are distributed unevenly across PV systems. FL is also utilized for detecting false data injection attacks in solar farms in \cite{AFederatedLearningFrameworkforDetectingFalseDataInjectionAttacksinSolarFarms}, in which the authors further emphasize that FL significantly minimizes the communication costs compared to a (privacy-violating) data sharing setting. All studies used \texttt{FedAvg} as aggregation algorithm.
\newline \newline

\noindent \textbf{Smart buildings.} The collected wealth of data from buildings opens up the opportunity to detect anomalous behavior based on data from smart meters. With FL, the clients, typically residential or office buildings, share sensitive energy consumption information in a privacy-preserving manner. The studies in \cite{UnsupervisedFederatedandPrivacy-PreservingDetectionofAnomalousElectricityConsumptioninReal-WorldScenarios} and \cite{AFederatedLearningApproachtoAnomalyDetectioninSmartBuildings} investigate the use of \texttt{FedAvg} for anomaly detection in smart buildings. The authors of \cite{UnsupervisedFederatedandPrivacy-PreservingDetectionofAnomalousElectricityConsumptioninReal-WorldScenarios} report significantly better results compared to local models, while \cite{AFederatedLearningApproachtoAnomalyDetectioninSmartBuildings} demonstrates improvements over centralized baseline models and also a faster convergence compared to a centralized benchmark. 

Both contributions employ additional privacy mechanisms, in \cite{AFederatedLearningApproachtoAnomalyDetectioninSmartBuildings} with secure multi-party computation and differential privacy, and through a privacy-preserving aggregation with shares in \cite{UnsupervisedFederatedandPrivacy-PreservingDetectionofAnomalousElectricityConsumptioninReal-WorldScenarios}.

\subsection{Federated control}
An increasing amount of information-rich client data is becoming available with the development of smart meters and other sensor-equipped systems monitoring energy consumption patterns and schedules in applications for, e.g., smart homes or electric vehicle fleets. Federated learning for control tasks based on this data can enable a more informed production, usage and storage of renewable energy, ranging from building energy management to vehicle fleet charging coordination. 
\newline  

\noindent \textbf{Smart meters.} A \texttt{FedSGD} based approach is presented in \cite{FederatedReinforcementLearningforEnergyManagementofMultipleSmartHomesWithDistributedEnergyResources} for the energy management of multiple smart homes, observing increased efficiency in terms of data storage and communication compared to a centralized approach. A \texttt{FedAvg}-based framework for thermal comfort control is explored in \cite{FederatedLearningforEnergy-efficientThermalComfortControlServiceinSmartBuildings}, demonstrating its effectiveness while further showing a positive impact of personalization as realized with a fine-tuning step. For electric vehicle fleet charging, an attention-weighted \texttt{FedAvg} technique is presented in \cite{AMultiagentFederatedReinforcementLearningApproachforPlug-InElectricFleetChargingCoordinationinaResidentialCommunity}, both outperforming a centralized benchmark model and the unweighted \texttt{FedAvg} approach. 

\subsection{Previous studies}
An overview of studies on federated learning in renewable energy applications is outlined in Table \ref{table}, highlighting the respective FL methods and the learning tasks accomplished by them.\\

	\setlength\LTleft{-1.6cm}\renewcommand{\arraystretch}{2.0}\begin{longtable}{ccllc}
		\hline
		\textbf{Reference} & \textbf{Centralized} & \textbf{Method} & \textbf{Task} & \makecell[c]{\textbf{Application} \\ \textbf{Field}}\\
		\hline 

% ---------- wind ---------

  		\cite{TowardsFleet-wideSharingofWindTurbineConditionInformationthroughPrivacy-preservingFederatedLearning}& \checkmark  & \makecell[l]{\texttt{FedAvg} with \\ fine-tuning}  & \makecell[l]{WT condition monitoring}  & W\\
		\hline % 11

  		\cite{Windpowerforecastingconsideringdataprivacyprotection} & \checkmark  & \makecell[l]{\texttt{FedAvg} for\\ reinforcement learning } & \makecell[l]{Wind power forecasting}  & W\\  
		\hline % 68

        \cite{DeepFederatedLearning-BasedPrivacy-PreservingWindPowerForecasting} & \checkmark & \makecell[l]{\texttt{FedAvg}} & \makecell[l]{Wind power forecasting}  & W\\
        \hline % 69

        \cite{ACyber-Securegeneralizedsupermodelforwindpowerforecastingbasedondeepfederatedlearningandimageprocessing} & \checkmark & \texttt{FedAvg} & \makecell[l]{Wind power forecasting}  & W\\
        \hline % 70

        \cite{APrivacy-preservingWindSpeedPredictionMethodBasedonFederatedDeepLearning} & 
 \checkmark & \texttt{FedProx} & \makecell[l]{Wind speed prediction}  & W\\
        \hline % 71

\cite{AnefficientfederatedtransferlearningframeworkforcollaborativemonitoringofwindturbinesinIoE-enabledwindfarms}&\checkmark &  \makecell[l]{ \texttt{FedAvg} with\\ client selection and \\domain adaptation} & \makecell[l]{WT anomaly detection \\  (blade cracking)} & W\\
        \hline % 72

  		\cite{DeepFedWT:Afederateddeeplearningframeworkforfaultdetectionofwindturbines} & \checkmark  & \texttt{FedAvg} & \makecell[l]{Fault detection in WTs} & W\\
		\hline % 88

		\cite{FederatedMulti-ModelTransferLearningBasedFaultDiagnosiswithPeer-to-PeerNetworkforWindTurbineCluster}& \checkmark  &  \makecell[l]{Clustered \texttt{FedAvg} with \\ dynamic weights} & \makecell[l]{Fault diagnosis in WTs}  & W\\
		\hline % 89

  		\cite{Windturbinebladeicingdetection:afederatedlearningapproach} & \checkmark  & \makecell[l]{\texttt{FedAvg}} & \makecell[l]{WT blade icing detection}  & W\\
		\hline % 90

         \cite{AClass-ImbalancedHeterogeneousFederatedLearningModelforDetectingIcingonWindTurbineBlades} & \checkmark  & \makecell[l]{FL with \\ exchanged feature maps} &  \makecell[l]{WT blade icing detection}  & W\\
        \hline % 91

         \cite{HumanKnowledge-basedCompressedFederatedLearningModelforWindTurbineBladeIcingDetection} & \checkmark  & \makecell[l]{\texttt{FedAvg} with\\ custom weights} & \makecell[l]{WT blade icing detection}  & W
 \\
        \hline % 92

        \cite{ABlockchain-EmpoweredCluster-BasedFederatedLearningModelforBladeIcingEstimationonIoT-EnabledWindTurbine}& \checkmark & \makecell[l]{Clustered \texttt{FedAvg}} & \makecell[l]{WT blade icing detection}  & W\\
        \hline % 93

%%% -------- SOLAR ------
        \cite{NovelPVPowerHybridPredictionModelBasedonFLCo-TrainingMethod} & \checkmark & \texttt{FedAvg} & \makecell[l]{PV power prediction}  & S\\
        \hline % 74

		\cite{Asolarforecastingframeworkbasedonfederatedlearninganddistributedcomputing} & \checkmark  & \makecell[l]{\texttt{FedAvg} with \\ personalization layers} & \makecell[l]{Solar forecasting} & S\\
		\hline     % 76   

		\cite{FuzzyClusteredFederatedLearningAlgorithmforSolarPowerGenerationForecasting}& \checkmark  & \makecell[l]{Fuzzy clustered FL} & \makecell[l]{Solar power \\ generation forecasting}  & S\\
		\hline % 77

        \cite{APrivacy-PreservingFederatedLearningMethodforProbabilisticCommunity-LevelBehind-the-MeterSolarGenerationDisaggregation} & \checkmark & \makecell[l]{\texttt{FedAvg} with \\ personalization layers}  & \makecell[l]{Community-level \\ behind-the-meter \\ solar generation disaggregation}  & S\\
        \hline % 78

        \cite{AsynchronousDecentralizedFederatedLearningforCollaborativeFaultDiagnosisofPVStations} & \checkmark & \makecell[l]{Asynchronous, \\ decentralized \texttt{FedAvg}} & \makecell[l]{PV fault diagnosis}  & S\\
        \hline % 94

\cite{CollaborativelyIGBTOpencircuitFaultsinPhotovoltaicInverters:ADecentralizedFederatedLearning-basedMethod} & \xmark & \makecell[l]{Decentralized \texttt{FedAvg}} & \makecell[l]{Diagnosing faults \\in PV inverters}  & S\\
        \hline % 95

        \cite{AFederatedLearningFrameworkforDetectingFalseDataInjectionAttacksinSolarFarms}&\checkmark & \makecell[l]{\texttt{FedAvg}} &  \makecell[l]{Detecting false data\\  injection  attacks \\ in solar farms}  & S\\
        \hline % 96

        \cite{APVFaultDiagnosisFrameworkBasedonAsynchronousFederatedLearning} & \checkmark & \makecell[l]{Asynchronous, \\ decentralized \texttt{FedAvg}}& \makecell[l]{PV fault diagnosis}  & S\\
        \hline % 102

% -------- BUILDINGS -------------

		\cite{PrivacyEnhancedEnergyPredictioninSmartBuildingusingFederatedLearning} & \checkmark  & \makecell[l]{\texttt{FedAvg}}  & \makecell[l]{Energy prediction in \\ smart buildings}  & B\\
		\hline % 26

         \cite{Privacy-preservingknowledgesharingforfewshotbuildingenergyprediction:Afederatedlearningapproach} & \checkmark & \makecell[l]{Clustered \texttt{FedAvg}} & \makecell[l]{Few-shot building \\energy prediction}  & B\\
        \hline    % 60                  

\cite{ImprovingtheAccuracyofLoadForecastingforCampusBuildingsBasedonFederatedLearning}  & \checkmark & \texttt{FedAvg} &\makecell[l]{Load forecasting for\\ campus buildings}  & B\\
        \hline % 79

        \cite{FederatedLearningforShort-TermResidentialLoadForecasting} & \checkmark & \makecell[l]{Clustered \texttt{FedAvg} \\ with fine-tuning} & \makecell[l]{Short-term residential \\ load forecasting}  & B\\
        \hline % 80
        
               \cite{PersonalizedFederatedDARTSforElectricityLoadForecastingofIndividualBuildings} & \checkmark & \makecell[l]{Clustered \texttt{FedAvg} \\ with fine-tuning} & \makecell[l]{Electricity load \\ forecasting\\  of buildings}  & B\\
        \hline % 81

		\cite{Federatedlearning-basedshort-termbuildingenergyconsumptionpredictionmethodforsolvingthedatasilosproblem}& \checkmark & \makecell[l]{\texttt{FedAvg} with \\ similar client selection \\ and fine-tuning} & \makecell[l]{Short-term load \\ forecasting}  & B\\
		\hline % 82

          \cite{ImpactofFederatedLearningOnSmartBuildings} & \xmark& \makecell[l]{Personalized \texttt{FedAvg}} &  \makecell[l]{FL for smart buildings}  & B\\
        \hline % 83
        
        		\cite{DeepFederatedAdaptation:AnAdaptativeResidentialLoadForecastingApproachwithFederatedLearning} & \checkmark  & \makecell[l]{\texttt{FedAvg} with \\ domain adaptation}  &  \makecell[l]{Residential short-term \\ load forecasting}  & B\\
		\hline % 84

        \cite{ASecureFederatedDeepLearning-BasedApproachforHeatingLoadDemandForecastinginBuildingEnvironment} & \checkmark & \makecell[l]{\texttt{FedAvg}} & \makecell[l]{Heating load \\ forecasting for buildings}  & B\\
        \hline % 85  
        \cite{Anadaptivefederatedlearningsystemforcommunitybuildingenergyloadforecastingandanomalyprediction} & \checkmark & \makecell[l]{\texttt{FedAvg} with \\ adaptive weights} &  \makecell[l]{Building energy  \\ load forecasting\\ and anomaly detection} & B\\
		\hline % 86

        \cite{DecentralizedFederatedLearningFrameworkfortheNeighborhood:ACaseStudyonResidentialBuildingLoadForecasting} & \xmark & \makecell[l]{Decentralized \texttt{FedAvg}}& \makecell[l]{Residential building \\ load forecasting}  & B\\
        \hline % 87

        \cite{UnsupervisedFederatedandPrivacy-PreservingDetectionofAnomalousElectricityConsumptioninReal-WorldScenarios}& \checkmark & \texttt{FedSGD} & \makecell[l]{Detection of anomalous \\ electricity consumption}  & B\\
        \hline % 97
     
        \cite{AFederatedLearningApproachtoAnomalyDetectioninSmartBuildings} & \checkmark & \makecell[l]{\texttt{FedAvg}}  & \makecell[l]{Anomaly detection for \\ buildings}  & B\\
        \hline % 98

\cite{FederatedReinforcementLearningforEnergyManagementofMultipleSmartHomesWithDistributedEnergyResources} & \checkmark & \makecell[l]{Federated reinforcement \\ learning} & \makecell[l]{Energy
management of \\ multiple smart homes}  & B\\
        \hline % 99

        \cite{FederatedLearningforEnergy-efficientThermalComfortControlServiceinSmartBuildings} & \checkmark & \makecell[l]{\texttt{FedAvg} with \\ fine-tuning} & \makecell[l]{Energy-efficient \\ thermal comfort \\ control service \\ in buildings}  & B\\
        \hline % 100
        \cite{Privacy-PreservingEnergyManagementofaSharedEnergyStorageSystemforSmartBuildings:AFederatedDeepReinforcementLearningApproach} & \checkmark & \makecell[l]{Federated reinforcement \\ learning} & \makecell[l]{Energy management \\ of shared energy storage}  & B\\
        \hline % 103

\cite{FederatedLearningEnabledPredictionofEnergyConsumptioninTransactiveEnergyCommunities}& \checkmark & \makecell[l]{\texttt{FedAvg}} & \makecell[l]{Prediction of energy \\consumption}  & B\\
        \hline % 104

       \cite{Federatedreinforcementlearningforsmartbuildingjointpeertopeerenergyandcarbonallowancetrading} & \checkmark  & \makecell[l]{\texttt{FedAvg} for \\ reinforcement learning} & \makecell[l]{Energy and carbon \\ allowance trading \\ for buildings}  & B\\
        \hline % 105
       \cite{FederatedOfficePlug-LoadIdentificationforBuildingManagementSystems} & \checkmark  & \texttt{FedAvg} & \makecell[l]{Plug-load identification \\ for building \\
management systems}  & B\\
        \hline % 106

        \cite{Privacy-PreservingRegulationCapacityEvaluationforHVACSystemsinHeterogeneousBuildingsBasedonFederatedLearningandTransferLearning} & \checkmark & \texttt{FedProx} & \makecell[l]{ Capacity evaluation for\\ HVAC  systems in \\ buildings}   & B\\
        \hline % 107

% ------ OTHER -----------------
        \cite{AMultiagentFederatedReinforcementLearningApproachforPlug-InElectricFleetChargingCoordinationinaResidentialCommunity} & \checkmark & \makecell[l]{Attention-weighted \\ \texttt{FedAvg}} & \makecell[l]{Plug-in electric vehicle \\ fleet charging coordination}  & O\\
        \hline

        \cite{AnIntelligentPrivacyPreservationSchemeforEVChargingInfrastructure}& \checkmark  & \texttt{FedAvg} & \makecell[l]{Electric vehicle charging}  & O  \\
        \hline

         \cite{FederatedLearningforimprovedpredictionoffailuresinAutonomousGuidedVehicles} &  \checkmark & \texttt{FedAvg} & \makecell[l]{Fault prediction \\in  autonomous \\guided vehicles}  & O\\
        \hline

	\hline

  \caption{Federated learning in data-driven renewable energy applications.\\ B: buildings, S: solar, W: wind, O: others.} % needs to go inside longtable environment
\label{table}
\end{longtable}

\section{Potential, challenges and future directions}
\label{sec:potential_challenges_and_futur_directions}

We will now analyze and discuss the potential and challenges of federated learning methods in the context of renewable energy. Moreover, we outline research needs and highlight possible future research directions. Our discussion in summarized in Table \ref{tbl:SummarySec5}.

\subsection{Potential of FL in renewable energy systems}
Previous studies have demonstrated that federated learning methods can be beneficial for overcoming data scarcity conditions as exemplified by \cite{Kusiak2016}. The results reported in previous studies consistently illustrate the superior ability of FL compared to local machine learning models in applications such as forecasting, control, and fault detection and diagnostics. They also demonstrated increased efficiency compared to centralized data sharing settings, and inherent generalization properties of models trained through FL. \\

Federated learning approaches seem particularly suited for renewable energy sectors that involve \emph{fleets} of distributed systems that generate, store, or consume energy, and that offer continuous sensor monitoring for the intended FL task. For example, fleets of wind turbines equipped with anemometers (to measure local wind conditions) and accelerometers (to measure vibration responses of critical drive train components) offer the potential to collaboratively learn power forecast models and fault detection models that outperform locally trained models, in a data-privacy-preserving manner based on FL. Similarly, \emph{fleets} of PV systems, smart meters, and vehicle batteries may also benefit from federated learning. The reviewed studies have demonstrated that FL can extract and share information collaboratively across the clients (fleet members) participating in the training. This can be achieved in a privacy-preserving manner where data remains local and inaccessible to other clients, effectively addressing the lack of data sharing and privacy concerns in this field. FL is also a highly flexible approach, shown to be compatible with a wide variety of data types, model architectures, and tasks commonly faced in renewable energy applications. Various proposed extensions to enhance security, efficiency, or model performance further demonstrate the option of adjusting FL methods to specific requirements. \\

There are numerous applications in the field of renewable energy in which the potential of federated learning has barely been explored yet. Examples include predictions of the remaining useful life of equipment, and estimating optimal maintenance schedules in predictive maintenance tasks. Furthermore, the vast majority of applications presented make use of \texttt{FedAvg} only with limited modifications and do not fully exploit recent state-of-the-art research contributions to further improve FL in practice, especially in terms of data heterogeneity and efficiency. Generally, the choice of framework, algorithms, and extensions is scattered in the presented literature, suggesting a research need for more in-depth comparisons of the applied FL algorithms across various application domains.\\

It is also apparent that some energy-related domains, such as electrical load forecasting of buildings, present a more advanced state of research of federated learning applications than others, such as battery management and control. When planning FL studies, it may be beneficial for researchers in renewable energy application domains less advanced in FL case studies to also take into account previous studies that were conducted on related tasks in other application domains with more comprehensive FL research, such as building load forecasting. \\

\begin{table}[h]
\begin{tabularx}{\linewidth}{>{\parskip1ex}X@{\kern4\tabcolsep}>{\parskip1ex}X}
\toprule
\hfil\bfseries \normalsize{\textbf{Advantages}}
&
\hfil\bfseries \normalsize{\textbf{Challenges and research needs}}
\\\cmidrule(r{3\tabcolsep}){1-1}\cmidrule(l{-\tabcolsep}){2-2}

\normalsize{+ FL enables collaborative learning while respecting data privacy\par
+ FL is well suited to small-scale distributed assets in fleets\par
+ Can consistently outperform local models\par
+ Highly flexible framework customizable to specific requirements\par
+ Customization techniques can address statistical heterogeneity\par
+ Privacy guarantees can be further enhanced\par} 

&

\normalsize{- Not reaching the performance of a data-sharing setting in all cases \par
- Scalability of FL to a larger number of client systems require further (field) testing\par
- Alternatives to \texttt{FedAvg} largely unexplored application-wise\par
- Chosen methods rarely justified and compared in studies \par
- Uneven progress across renewable energy application domains\par
- Non-federated hyper-parameter tuning\par
- Fairness and biases aspects have barely been investigated yet\par}

\\\bottomrule
  
\end{tabularx}
\caption{Summary of our findings.}
\label{tbl:SummarySec5}
\end{table}

\subsection{FL challenges and future directions}
Challenges, open questions and potential future research directions related to federated learning in renewable energy will be discussed in the following. 

The vast majority of studies reviewed employ a standard network configuration based on a centralized server, coupled with a relatively small number of clients. Uncertainties regarding the \textbf{scalability} of the proposed FL method are occasionally expressed across literature, especially concerning scenarios with a large number of participating clients. Only few contributions propose alternative approaches based on configurations with asynchronous, decentralized (server-less), and/or peer-to-peer network structures. In all of these cases, a substantial reduction in communication costs was successfully demonstrated, providing an essential finding which has not been sufficiently considered yet in other studies. Moreover, possible effects of the computational capabilities of clients and their differences on the learning process have not been investigated. The positive results of alternative network configurations provide promising research directions towards a more scalable and efficient use of FL in practical renewable energy applications. Further large-scale studies, based on e.g., more clients and realistic computational capabilities, are needed to confirm the effectiveness of the proposed alternatives. \\

\textbf{Statistical heterogeneity}, i.e. non-iid data, is a re-occurring major challenge in all application fields in renewable energy. Several studies report that non-iid data, if left unaddressed, negates any performance benefits of applying standard FL frameworks in their case studies, resulting in up to substantially inferior models than locally trained models. Personalized FL is dominantly used with significant success to mitigate this issue. Numerous techniques, mostly based on clustering, personalization layers, and fine-tuning, manage to outperform localized, non-personalized, and even centralized approaches in all presented cases. 

Thus, personalization appears to be a viable solution to mitigate non-iid data issues in renewable energy applications. However, existing studies usually propose only one specific personalization method, lacking its justification and comparisons with other personalization methods. While no personalization method is consistently superior in all applications (section \ref{sec:pers_FL}), comparisons may still be worthwhile and reveal crucial insights on differences across methods. Furthermore, the extent of distribution differences varies strongly across applications (e.g., individual household consumption vs. power generation of a fleet of PV systems), such that an appropriate choice of personalization method (if any) is challenging. It should be noted that this choice may have broader impacts. For instance, clustering or personalization layers affect the cold-start problem, as no single global model is available to distribute to newly joined clients. Additionally, some methods require changes to the framework which require consent of all participants, unlike, for example, fine-tuning which is essentially an optional, local post-FL step.

\texttt{FedAvg} is still predominantly used despite these non-iid issues. Alternative algorithms such as \texttt{FedProx} and \texttt{FUALA}, which aim to improve convergence and performance with non-iid data, remain largely unused in renewable energy applications. Overall, these results suggest that personalized FL is a very promising and viable solution for renewable energy applications even when faced with significant statistical heterogeneity. The potential benefits of applying alternative aggregation algorithms should, however, be investigated, given the prevalence of this issue. \\

Several studies have proposed weighted variants of \texttt{FedAvg} motivated by ensuring fairness. However, multiple challenges related to systematic \textbf{biases and fairness} in renewable energy applications remain unaddressed. For example, the impacts of an over-representation of e.g., systems operated only under certain weather conditions are largely unexplored across applications. Case studies are also lacking with regard to model robustness. Case studies investigating potential performance impacts caused by faulty sensor data are required, also with a view towards practical implementations of FL, yet remain absent. The potential of frameworks such as Agnostic Federated Learning addressing these challenges have also yet to be investigated in renewable energy applications.\\

Some stakeholders may seek more data \textbf{privacy} guarantees than already provided by a standard FL framework. This is reflected in the literature with numerous works employing additional privacy mechanisms. A wide variety of proposed enhancements in various combinations is shown to be compatible with FL for renewable energy applications, independently of the task and dataset. Further studies should investigate comparisons of privacy methods in terms of efficiency and their impact on model performance to support decision makers in justifying and choosing appropriate privacy solutions. \\

\noindent \textbf{Framework considerations.} The federated learning framework offers flexibility in its selection of the employed aggregation algorithm, communication structure, client selection, or the client training procedure. While numerous different frameworks and modifications of FL have been proposed across renewable energy application domains, a comprehensive comparison is lacking. As the optimal choice of a framework may depend on the application domain and data, a more comprehensive understanding of the performance of these different methods would support decisions of choosing an appropriate framework. \\

Regarding the choice of the aggregation, the algorithms discussed in our review and more generally applied in the field of renewable energy are all based on neuron-wise averaging, i.e., they aggregate and average the weights unit-by-unit for each neural network layer. Approaches of federated learning with dynamical network architectures such as Federated Match Averaging (\texttt{FedMA}, \cite{FederatedLearningwithMatchedAveraging}), Probabilistic Federated Neural Matching (\texttt{PFNM}, \cite{ProbabilisticFederatedNeuralMatching}) or Federated Distance (\texttt{FedDist}, \cite{Afederatedlearningaggregationalgorithmforpervasivecomputing:Evaluationandcomparison}) account for the various roles each unit plays in neural networks. Federated learning is still a rapidly growing field with new proposed modifications to improve efficiency or performance, and yet recent state-of-the-art federated learning algorithms have remained unused for renewable energy applications so far (including \texttt{FedMA} and \texttt{PFNM}, as mentioned above). These algorithms have demonstrated improvements in    performance, a reduction in required communication, and they can facilitate the choice of model architecture. Future research studies should investigate if further benefits can be obtained by applying recent state-of-the art FL methods, such as \texttt{FedMA}, \texttt{PFNM}, and others. Moreover, an adaptation of transfer learning methods in a federated learning setting might improve the issue of data heterogeneity and can also be subject of future research.\\

\noindent \textbf{Hyper-parameter tuning.} Another challenge in federated learning is the choice of model architecture and hyper-parameter tuning. In general, model architecture and parameter tuning are not dealt with in a federated way, meaning that the appropriate hyper-parameters are usually guessed from observing a single device, such as a single WT \cite{TowardsFleet-wideSharingofWindTurbineConditionInformationthroughPrivacy-preservingFederatedLearning}. But decisions based on single local clients may not represent the best global choice. For example, a neural network architecture may easily overfit on local clients with fewer, simpler, or narrowly distributed data, but may underfit the global task across clients. To address this problem, a federated approach to hyper-parameter tuning (\cite{DemystifyingHyperparameterOptimizationinFederatedLearning, Federatedhyperparametertuning:Challengesbaselinesandconnectionstoweight-sharing}) should be considered and investigated in the renewable energy applications of interest.
\newline \newline

\textbf{Implementation challenges.} Processes and policies related to federated learning still need to be defined to enable its implementation in practice. This includes specifying the roles and responsibilities of all actors related to the FL process. Different configurations and role distributions are possible in practice. For example, the federated learning process may be orchestrated by a regulatory entity, which might also define the neural network structure. This regulator could decide on what happens if a client disagrees with aspects of the federated training, for example, the client might be exempt from the training process or required to participate anyways. Federated learning processes can also be implemented and orchestrated by operators to enable data access across the fleet. Federated learning can, in principle, even be implemented by the manufacturer for customers who prefer not to give the manufacturer access to their data. Many research needs and open questions remain with regard to the actual practical implementations of federated learning in renewable energy systems.

\section{Conclusions}
\label{sec:conclusion}

With the growing importance of renewable energy in the energy mix, more accurate forecasting, condition monitoring, and control capabilities become crucial. Data-driven models based on deep learning have shown impressive results. However, these models require large and representative training datasets, which are often lacking. While utilizing data from other fleet members within and across fleets can be a solution, it is usually hindered by a lack of data sharing. Federated learning emerged as an approach for distributed systems to collaboratively train machine learning models in a data-privacy-preserving manner. It fully addresses data scarcity and privacy concerns as the locally stored datasets of each participant remain inaccessible to other participants. \\

Numerous studies have investigated potential applications of federated learning in the context of renewable energy generation, consumption and storage. We have provided a comprehensive review of the findings of those studies. Further, we have discussed challenges relevant to federated learning in renewable energy applications. Our analysis shows that FL can indeed enable privacy-preserving learning and consistently improve models trained on single asset datasets across all renewable energy application fields. As FL incorporates information from multiple assets, participants benefit from e.g., more accurate forecasting (by learning from various geographical locations), improved fault detection (by obtaining a larger fault database), and improved asset control (by learning from consumption patterns of multiple assets). We further note that the flexibility of federated learning approaches allowed researchers to develop approaches customized to specific requirements, e.g., with regard to data heterogeneity or additional privacy needs. Our findings suggest a large potential of federated learning in all considered renewable energy applications. \\

However, many challenges and research questions remain open. One challenge relates to the scalability  of FL to larger fleets, which requires further investigation. The robustness and fairness of FL also remain unexplored due to a lack of exploration into potential biases or, e.g., effects of faulty sensor data on the FL performance. While previous studies demonstrated numerous privacy enhancements, a more unified approach with in-depth comparisons is required to support decision makers in choosing appropriate solutions. Statistical heterogeneity has appeared as a major prevalent issue across applications, due to commonly arising distribution differences between client data, such as differences in the technical specifications of the power plant or of individual energy consumption patterns. We find that model customization approaches can successfully mitigate model performance degradation resulting from statistical heterogeneity. Customization techniques have been demonstrated in some cases to be essential when non-customized FL methods are outperformed by locally trained models.\\

Despite the prevalence of these challenges, alternative aggregation algorithms to \texttt{FedAvg} which could address these open questions have received insufficient attention.  If further research addresses the presented challenges, federated learning may help overcome the lack of data sharing and, thereby, enable more innovation and efficiency in renewable energy applications.

\section*{Acknowledgements}
This research was supported by the Swiss National Science Foundation (Grant No. 206342). 

\bibliography{biblio2}

\end{document}